\documentclass{sig-alternate-05-2015}

\newif\ifapx
\apxtrue 

\usepackage{times}
\usepackage[english]{babel}

\makeatletter
\newif\if@restonecol
\makeatother

\usepackage{graphicx}
\usepackage{algorithm}
\usepackage{algpseudocode}

\usepackage{amsmath}
\usepackage{amssymb}
\usepackage{color}

\usepackage{balance}
\usepackage{multirow}
\usepackage{verbatim}
\usepackage{fixltx2e}
\usepackage{bm}
\usepackage{url}

\usepackage{eqparbox}

\usepackage{booktabs}  
\usepackage{subfig}

\usepackage{xspace}
\usepackage{microtype}

\usepackage{mathtools}
\mathtoolsset{showonlyrefs=true}

\usepackage{tikz}
\usetikzlibrary{decorations,arrows,shapes,positioning,fit,calc,plotmarks}
\usepackage{pgfplots}
\usepackage{ifpdf}

\ifpdf
\usetikzlibrary{external}
\fi

\newcommand{\algname}{Ditto}
\newcommand{\ourmethod}{\textsc{\algname}\xspace}

\newcommand{\Sax}{\textsc{sax}\xspace}
\newcommand{\Sqs}{\textsc{Sqs}\xspace}

\newcommand{\oururl}{\url{http://eda.mmci.uni-saarland.de/ditto/}}

\algtext*{EndWhile}
\algtext*{EndIf}
\algtext*{EndFor}

\renewcommand{\algorithmiccomment}[1]{\bgroup\hfill\scriptsize//~#1\egroup}

\algnewcommand\algorithmicinput{\textbf{Input:}}
\algnewcommand\INPUT{\item[\algorithmicinput]}

\algnewcommand\algorithmicoutput{\textbf{Output:}}
\algnewcommand\OUTPUT{\item[\algorithmicoutput]}

\renewcommand{\S}{S}	
\newcommand{\occs}{\mathit{occs}}	
\newcommand{\CS}{\mathit{PS}}	
\newcommand{\C}{\mathcal{C}}	

\newcommand{\Cp}{C_p}	
\newcommand{\Cg}{C_g}	

\newcommand{\codepattern}{\code_\mathit{p}}
\newcommand{\codegap}{\code_\mathit{g}}
\newcommand{\codefill}{\code_\mathit{f}}

\newcommand{\LN}{L_\mathbb{N}}

\newlength{\tilelen} 
\newcommand{\synhead}{
	\begin{tabular}{rrrr c rrr c rrrr c rr} 
	\toprule
	\multicolumn{4}{l}{\textbf{Data}} && \multicolumn{3}{l}{\textbf{Planted Patterns}} && \multicolumn{7}{l}{\textbf{Discovered Patterns}} \\
	\cmidrule{1-4}
	\cmidrule{6-8}
	\cmidrule{10-16}					
	&&&&&&&&&&&& && \multicolumn{2}{r}{\textbf{Time (sec)}} \\
	\cmidrule{15-16}
	$||D||$ & $t(\DB)$ & $|A|$ & $|\Omega_i|$ && $|\P|$ & $||X||$ & $\support$ && $=$ & $\subset$ & $\cup$ & ? && $mean$ & $std$  \\
	\midrule	
}
\newcommand{\synfoot}{\end{tabular}}

\newcommand{\mev}[2]{\genfrac{}{}{0pt}{}{#1}{#2}}
\newcommand{\cand}{\mathit{Cand}}
\newcommand{\gaps}{\mathit{gaps}}
\newcommand{\usage}{\mathit{usage}}
\newcommand{\support}{\mathit{support}}

\newcommand{\fills}{\mathit{fills}}
\newcommand{\code}{\mathit{code}}

\makeatletter
\renewcommand*{\@fnsymbol}[1]{\ensuremath{\ifcase#1\or \circ\or \bullet\or \ddagger\or
   \mathsection\or \mathparagraph\or \|\or **\or \dagger\dagger
   \or \ddagger\ddagger \else\@ctrerr\fi}}
\makeatother

\newcommand{\CT}{\mathit{CT}}
\newcommand{\ST}{\mathit{ST}}

\newcommand{\DB}{D}
\renewcommand{\P}{\mathcal{P}}

\newcommand{\Model}{M}
\newcommand{\Models}{\mathcal{M}}

\xspaceaddexceptions{$}

\makeatletter 
\pgfdeclareshape{event}{ 
\inheritsavedanchors[from=rectangle] 
\inheritanchorborder[from=rectangle] 
\inheritanchor[from=rectangle]{center} 
\inheritanchor[from=rectangle]{north} 
\inheritanchor[from=rectangle]{south} 
\inheritanchor[from=rectangle]{south east} 
\inheritanchor[from=rectangle]{south west} 
\inheritanchor[from=rectangle]{west} 
\inheritanchor[from=rectangle]{east} 
\backgroundpath{
\southwest \pgf@xa=\pgf@x \pgf@ya=\pgf@y 
\northeast \pgf@xb=\pgf@x \pgf@yb=\pgf@y 
\pgf@yc=\pgf@ya \advance\pgf@yc by 2pt 
\pgfpathmoveto{\pgfpoint{\pgf@xa}{\pgf@yc}} 
\pgfpathlineto{\pgfpoint{\pgf@xa}{\pgf@ya}} 
\pgfpathlineto{\pgfpoint{\pgf@xb}{\pgf@ya}} 
\pgfpathlineto{\pgfpoint{\pgf@xb}{\pgf@yc}} 
} 
}
\makeatother

\tikzstyle{tile} = [rounded corners = 2pt, inner sep = 0pt, fill opacity = 0.3, anchor = south west, minimum width = 11pt, minimum height = 7pt]

\pgfdeclarelayer{background}
\pgfdeclarelayer{foreground}
\pgfsetlayers{background,main,foreground}

\tikzstyle{block} = [rounded corners, draw=blue!70, fill=white, text width=3.3cm, minimum height=4em]
\tikzstyle{bgblock} = [rounded corners, draw=blue!70, thick, fill=blue!10, text width=3.3cm, minimum height=4em]
\tikzstyle{line} = [draw, -latex', thick,blue!70]

\definecolor{yafaxiscolor}{rgb}{0.3, 0.3, 0.3}

\definecolor{yafcolor1}{rgb}{0.6, 0.016, 0.553}		
\definecolor{yafcolor2}{rgb}{0.7, 0.3, 0.0}			
\definecolor{yafcolor3}{rgb}{0.1, 0.6, 0.1}			
\definecolor{yafcolor4}{rgb}{0.01, 0.01, 0.01}		
\definecolor{yafcolor5}{rgb}{0.141, 0.345, 0.643}	
\definecolor{yafcolor6}{rgb}{0.4, 0.4, 0.0}			
\definecolor{yafcolor7}{rgb}{0.0, 0.3, 0.3}			
\definecolor{yafcolor8}{rgb}{0.925, 0.165, 0.224}	

\newlength{\yafaxispad}
\newlength{\yaftlpad}
\newlength{\yaflabelpad}
\newlength{\yafaxiswidth}
\newlength{\yafticklen}
\makeatletter
\def\pgfplots@drawtickgridlines@INSTALLCLIP@onorientedsurf#1{}
\makeatother

\newcommand{\yafdrawaxis}[4]{
	\pgfplotstransformcoordinatex{#1}\let\xmincoord=\pgfmathresult 
	\pgfplotstransformcoordinatex{#2}\let\xmaxcoord=\pgfmathresult 
	\pgfplotstransformcoordinatey{#3}\let\ymincoord=\pgfmathresult 
	\pgfplotstransformcoordinatey{#4}\let\ymaxcoord=\pgfmathresult 
	\pgfsetlinewidth{\yafaxiswidth} 
	\pgfsetcolor{yafaxiscolor}
	\pgfpathmoveto{\pgfpointadd{\pgfpointadd{\pgfplotspointrelaxisxy{0}{0}}{\pgfqpointxy{\xmincoord}{0}}}{\pgfqpoint{-0.5\yafaxiswidth}{\yafaxispad}}}
	\pgfpathlineto{\pgfpointadd{\pgfpointadd{\pgfplotspointrelaxisxy{0}{0}}{\pgfqpointxy{\xmaxcoord}{0}}}{\pgfqpoint{0.5\yafaxiswidth}{\yafaxispad}}}
	\pgfpathmoveto{\pgfpointadd{\pgfpointadd{\pgfplotspointrelaxisxy{0}{0}}{\pgfqpointxy{0}{\ymincoord}}}{\pgfqpoint{\yafaxispad}{-0.5\yafaxiswidth}}}
	\pgfpathlineto{\pgfpointadd{\pgfpointadd{\pgfplotspointrelaxisxy{0}{0}}{\pgfqpointxy{0}{\ymaxcoord}}}{\pgfqpoint{\yafaxispad}{0.5\yafaxiswidth}}}
	\pgfusepath{stroke}
}

\pgfplotscreateplotcyclelist{yaf}{%
{yafcolor1,mark options={scale=0.75},mark=o}, 
{yafcolor2,mark options={scale=0.75},mark=square},
{yafcolor3,mark options={scale=0.75},mark=triangle},
{yafcolor4,mark options={scale=0.75},mark=o},
{yafcolor5,mark options={scale=0.75},mark=o},
{yafcolor6,mark options={scale=0.75},mark=o},
{yafcolor7,mark options={scale=0.75},mark=o},
{yafcolor8,mark options={scale=0.75},mark=o}} 

\pgfplotsset{axis y line=left, axis x line=bottom,
	tick align=outside,
	tickwidth=\yafticklen,
	clip = false,
    x axis line style= {-, line width = 0pt, color=black!0},
    y axis line style= {-, line width = 0pt, color=black!0},
    x tick style= {line width = \yafaxiswidth, color=yafaxiscolor, yshift = \yafaxispad},
    y tick style= {line width = \yafaxiswidth, color=yafaxiscolor, xshift = \yafaxispad},
    x tick label style = {font=\scriptsize, yshift = \yaftlpad},
    y tick label style = {font=\scriptsize, xshift = \yaftlpad},
    every axis y label/.style = {at = {(ticklabel cs:0.5)}, rotate=90, anchor=center, font=\scriptsize, yshift = -\yaflabelpad},
    every axis x label/.style = {at = {(ticklabel cs:0.5)}, anchor=center, font=\scriptsize, yshift = \yaflabelpad},
    x tick label style = {font=\scriptsize, yshift = 1pt},
    grid = major,
    major grid style  = {dash pattern = on 1pt off 3 pt},
	every axis plot post/.append style= {line width=\yafaxiswidth} ,
	legend cell align = left,
	legend style = {inner sep = 1pt, cells = {font=\scriptsize}},
	legend image code/.code={%
		\draw[mark repeat=2,mark phase=2,#1] 
		plot coordinates { (0cm,0cm) (0.15cm,0cm) (0.3cm,0cm) };%
	} 
}

\begin{document}

\title{Keeping it Short and Simple:\\Summarising Complex Event Sequences with\\ Multivariate Patterns}

\numberofauthors{3} 
\author{
\alignauthor
Roel Bertens\\ 
       \affaddr{Department of Information and Computing Sciences}\\
       \affaddr{Utrecht University}\\
       \affaddr{Utrecht, The Netherlands}\\
       \email{R.Bertens@uu.nl}
\alignauthor
Jilles Vreeken\\ 
       \affaddr{Max Planck Institute for Informatics and\\ Saarland University}\\
       \affaddr{Saarbr\"{u}cken, Germany}\\       
       \email{jilles@mpi-inf.mpg.de}
\alignauthor 
Arno Siebes\\ 
       \affaddr{Department of Information and Computing Sciences}\\
       \affaddr{Utrecht University}\\
       \affaddr{Utrecht, The Netherlands}\\
       \email{A.P.J.M.Siebes@uu.nl}
}

\date{}

\maketitle

\begin{abstract} 

	We study how to obtain concise descriptions of discrete multivariate sequential data. In particular, how to do so in terms of rich multivariate sequential patterns that can capture potentially highly interesting (cor)relations between  sequences. To this end we allow our pattern language to span over the domains (alphabets) of all sequences, allow patterns to overlap temporally, as well as allow for gaps in their occurrences. 
	
	We formalise our goal by the Minimum Description Length principle, by which our objective is to discover the set of patterns that provides the most succinct description of the data. To discover high-quality pattern sets directly from data, we introduce \ourmethod, a highly efficient algorithm that approximates the ideal result very well. 
		
	Experiments show that \ourmethod correctly discovers the patterns planted in synthetic data. Moreover, it scales favourably with the length of the data, the number of attributes, the alphabet sizes. On real data, ranging from sensor networks to annotated text, \ourmethod discovers easily interpretable summaries that provide clear insight in both the univariate and multivariate structure. 
	
\end{abstract}	

\section{Introduction} \label{sec:intro}

	Most real sequential data is multivariate, as when we measure data over time, we typically do so over multiple attributes. 
	Examples include sensors in a network, frequency bands in seismic or ECG data, transaction records, and annotated text. In this paper we investigate how to succinctly summarise such data in terms of those patterns that are most characteristic for the data.
	
	To capture these characteristics, our pattern language needs to be rich, i.e., patterns may span multiple aligned sequences (attributes) in which no order between these attributes is assumed and occurrences may contain gaps. For example, if we consider a sensor network, a pattern could be a characteristic set of values for multiple attributes at one point in time, or, more complex, a specific value for one sensor, temporally followed by certain readings of one or more other sensors. That is, patterns that are able to identify associations and correlations between one or multiple attributes.
	
	Having such a rich pattern language has as immediate consequence that the well-known frequent pattern explosion hits particularly hard. Even for trivial data sets one easily finds enormous numbers of frequent patterns, even at modest minimal frequency thresholds~\cite{tatti:11:multievent}; the simplest synthetic dataset we consider contains only five true patterns, yet the \emph{lower bound} on the number of frequent patterns is more than a billion. 
	
	Clearly, returning collections of such magnitude is useless, as they cannot be inspected in any meaningful way. Rather, we want a small set of such patterns that collectively describe the data well. To find such sets, we employ the Minimum Description Length (MDL) principle \cite{grunwald:07:book}. This approach has a proven track record in the domain of transaction data mining \cite{vreeken:11:krimp,smets:12:slim} and has also been successfully applied to, e.g.\ sequential data \cite{tatti:12:sqs,lam:12:gokrimp}. Because complex multivariate sequential data is ubiquitous and correlation between multiple sequences can give much insight to domain experts, here we take this research further by defining a framework to summarise this data, while able to efficiently deal with the enormous space of frequent patterns. Note that, all real-valued time series can easily be discretised to fit our framework, for example using SAX~\cite{lin:07:sax}.

	The humongous number of frequent patterns makes it intractable to discover a small set of characteristic patterns by post-processing the set of all frequent patterns; all the more because the MDL objective function on pattern sets is neither monotone nor sub-modular. Hence we introduce a heuristic algorithm, called \ourmethod,\!\footnote{The ditto mark (") is used in lists to indicate that an item is repeated, i.e., a multivariate pattern.} that mines characteristic sets of patterns directly from the data. In a nutshell, \ourmethod mines good models by iteratively adding those patterns to the model that maximally reduce redundancy. That is, it iteratively considers the current description of the data and searches for patterns that most frequently occur one after the other, and considers the union of such pairs as candidates to add to the model. As such, \ourmethod focusses on exactly those patterns that are likely to improve our score, pruning large parts of the search space, and hence only needs seconds to mine high-quality summaries.
	
	Our extensive empirical evaluation of \ourmethod shows it ably discovers succinct summaries that contain the key patterns of the data at hand, and, importantly, that it does not pick up on noise. An example of what we discover includes the following. When applied to the novel Moby Dick by Herman Melville, on the topic of whaling, the summary \ourmethod discovers consists of meaningful non-trivial linguistic patterns including ``$\langle \text{noun} \rangle \text{ of the } \langle \text{noun} \rangle$", e.g.\ ``Man of the World'', and ``the $\langle \text{noun} \rangle$ of", as used in "the ship of (somebody)". These summaries hence give clear insight in the writing style of the author(s). Besides giving direct insight, the summaries we discover have high downstream potential. One could, for example, use them for comparative analysis~\cite{budhathoki:15:diffnorm}. For example, for text identifying similarities and differences between authors, for sensor networks detecting and describing concept drift over time, and characterising differences between patients. Such analysis is left for future work, as we first have to be able to efficiently discover high quality pattern-based summaries from multivariate sequential data. That is what we focus on in this paper.

The remainder of this paper is organised as follows. We first cover preliminaries and introduce notation in Section~\ref{sec:pre}. In Section~\ref{sec:the} we formally introduce the problem. Section~\ref{sec:alg} gives the details of the \ourmethod algorithm. We discuss related work in Section~\ref{sec:rel}, and empirically evaluate our score in Section~\ref{sec:exp}. We round up with discussion and conclusions in Sections~\ref{sec:dis} and \ref{sec:con}, respectively. 

\section{Preliminaries and Notation} \label{sec:pre}

		As data type we consider databases $\DB$ of $|\DB|$ multivariate event sequences $S \in \DB$, all over attributes $A$.
		We assume that the set of attributes is indexed, such that $A_i$ refers to the $i^{th}$ attribute of $D$. 
		
		A multivariate, or \emph{complex}, event sequence $S$ is a bag of $|A|$ univariate event sequences, $S = \{S_1, \ldots, S_{|A|}\}$. An event sequence $S_i \in \Omega_i^n$ is simply a sequence of $n$ events drawn from discrete domain $\Omega_i$, which we will refer to as its \emph{alphabet}. An event is hence simply an attribute--value pair. That is, the $j^{\mathit{th}}$ event of $S^i$ corresponds to the value of attribute $A_i$ at time $j$. We will write $\Omega$ for the union over these attribute alphabets, i.e.\ $\Omega = \displaystyle \cup_{i \in |A|} \Omega_i$.
		
		By $||S||$ we indicate the number of events in a multivariate sequence $S$, and by $t(S)$ the length of the sequence, i.e.\ the number of time steps. We will refer to the set of events at a single time step $j$ as a \emph{multi-event}, writing $S[j]$ for the $j^{\mathit{th}}$ multi-event of $S$. To project a multivariate event sequence $S$ onto an individual attribute, we write $S^j$ for the univariate event sequence on the $j^{th}$ attribute, and analogously define $D^j$. For completeness, we define $||\DB|| = \sum_{S \in \DB} ||S||$ for the total number of events, and $t(\DB) = \sum_{S} t(S)$ for the total number of time steps in $\DB$. 

		Our framework fits both categorical and transaction (item set) data. With categorical data the number of events at each time step is equal to the number of attributes. 
		For simplicity, w.l.o.g., we consider categorical data in the remainder.

\begin{figure}[h]
			\begin{center}
				\includegraphics[width=0.48\textwidth]{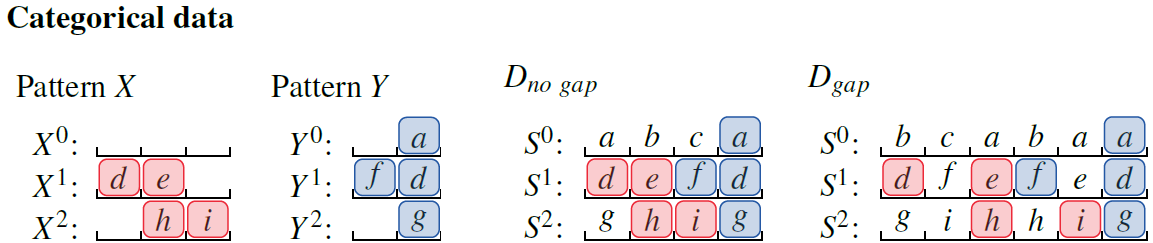}
			\end{center}
			\caption{Patterns $X$ and $Y$ occurring with ($\DB_{\mathit{gap}}$) and without ($\DB_{\mathit{no\ gap}}$) gaps in the data.}
			\label{fig:pat}
		\end{figure}

		As patterns we consider partial orders of multi-events, i.e. sequences of multi-events, allowing for gaps between time steps in their occurrences. By $t(X)$ we denote the \emph{length} of a pattern $X$, i.e.\ the number of time steps for which a pattern $X$ defines a value. 
		In addition, we write $||X||$ to denote the \emph{size} of a pattern $X$, the total number of values it defines, i.e.\ $ \sum_{X[i] \in X} \sum_{x \in X[i]} 1$. For example, in Figure~\ref{fig:pat} it holds that $||X|| = 4$ and $t(X) = 3$.
			
		We say a pattern $X$ is \emph{present} in a sequence $S \in \DB$ of the data when there exists an interval $[t_{start}, \cdots, t_{end}]$ in which all multi-events $X[i] \in X$ are contained in the order specified by $X$, allowing for gaps in time. That is, $\forall_{X[i] \in X} \exists_{j \in [start,end]} X[i] \subseteq S[j]$, and $k \ge j+1$ for $X[i] \subseteq S[j]$ and $X[i+1] \subseteq S[k]$. A singleton pattern is a single event $e \in \Omega$ and $\P$ is the set of all non-singleton patterns. A minimal window for a pattern is an interval in the data in which a pattern occurs which can not be shortened while still containing the whole pattern~\cite{tatti:09:signifeps}. In the remainder of this paper when we use the term pattern occurrence we always expect it to be a minimal window. 
		Further, we use $\occs(X,S)$ for the disjoint set of all occurrences of $X$ in $S$. 
		Figure~\ref{fig:pat} shows examples of two categorical patterns occurring both with and without gaps. 

		Our method operates on categorical data. To find patterns in continuous real-valued time series we first have to discretise it. \Sax \cite{lin:07:sax} is a celebrated approach for doing so, and we will use it in our experiments -- though we note that any discretisation scheme can be employed. This ordinal data can either be represented by absolute values per time step, or by relative values that represent the difference between each pair of subsequent values. These relative values describe the changes in the data rather than the exact values -- by which different types of patterns can be discovered. 

\section{Theory} \label{sec:the}	

	In this section we give a brief introduction to the MDL principle, we define our objective function, and analyse the complexity of the optimization problem.

	\subsection{MDL Introduction}\label{sec:the:mdl}
		The Minimum Description Length principle (MDL) \cite{grunwald:07:book} is a method for inductive inference that aims to find the best model for the data. It is based on the insight that any regularity in the data can be used to compress the data and the more we can compress, the more regularity we have found. More formally, given a set of models $\Models$, the best model $\Model \in \Models$ is the one that minimises 
		\[
			L(\Model) + L(\DB \mid \Model)
		\]
		where $L(\Model)$ is the length of the description of $\Model$, and $L(\DB \mid \Model)$ is the length of the description of the data. 
		
		To use MDL, we have to define how our model looks and how we can use it to describe the data. That is, we need to be able to encode and decode the data using our model.

	\subsection{MDL for Multivariate Sequential Data}	\label{sec:the:score}
		As models for our data we consider code tables ($\CT$) \cite{vreeken:11:krimp,tatti:12:sqs}; these are simple four-column look-up tables (or, dictionaries) between patterns on the left hand side, and associated codes on the right hand side. In this paper we consider three types of codes, i.e.\ pattern codes, gap codes and fill codes. We will explain the use of these below, in Section~\ref{subsec:enc} and Example~2. 
		
		Loosely speaking, whenever we read a pattern code $\codepattern(X \mid \CT)$ we know we can start to decode the events of pattern $X$. A fill code $\codefill(X \mid \CT)$ tells us we can decode the next time step of pattern $X$, whereas a gap code $\codegap(X \mid \CT)$ tells us there is a gap in this occurrence of pattern $X$. To fill such a gap we read the next pattern code. Note that, our approach thus allows for patterns to interleave.	For readability, we do not write $\CT$ wherever clear from context.		
		To make sure we can encode any multivariate event sequence over $\Omega$, we require that a code table at least contains all singleton events. 
		Next, we will describe how to use these code tables to cover and encode a dataset. Further, we define how to compute $L(\CT)$ and $L(\DB \mid \CT)$ and we conclude with our formal problem definition.
		
		\subsubsection{Covering} \label{subsec:cover}
			A cover determines where we use each pattern when encoding a dataset and also where gaps in patterns occur. More formally, a cover $\C$ defines the set of all pattern occurrences that are used to cover the complete dataset. Therefore, if one of the events from pattern occurrence $o$ is already covered by another pattern occurrence $o' \in \C$ we say that the occurrence $o$ \emph{overlaps} with the current cover $\C$, i.e. $\C \cap o \neq \emptyset$, which we do not allow. 
			
			As we are after the minimal description, we want to use optimal codes. Clearly, the more often a code is used, the shorter it should be. This, however, creates a non-trivial connection between how often a pattern is used to describe part of the data, and its code length. This, in turn, makes the complete optimisation process, and in particular that of finding a good cover of the data given a set of patterns non-trivial. For the univariate case, we showed~\cite{tatti:12:sqs} that when we know the optimal code lengths we can determine the optimal cover by dynamic programming, and that we can approximate the optimal cover by iteratively optimising the code lengths and the cover. Although computationally expensive, for univariate data this strategy works well. For multivariate data, naively seen the search space grows enormously. More importantly, it is not clear how this strategy can be generalised, as now the usages of two patterns are allowed to temporally overlap as long as they do not cover the same events. We therefore take a more general and faster greedy strategy.

\begin{figure}[t]
				\begin{center}
					\includegraphics[width=0.48\textwidth]{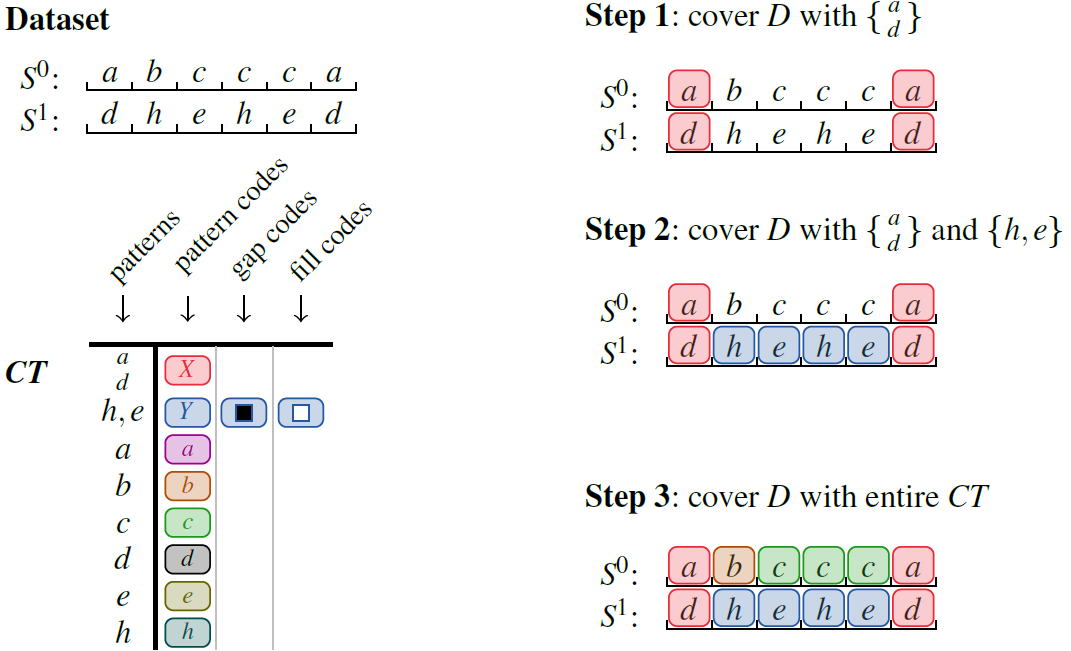}
				\end{center}
				\caption{Sample data $\DB$, code table $\CT$, and cover $\C$. 
				}
				\label{fig:toy1}
			\end{figure}

			To cover a dataset we need the set of patterns from the first column of a code table $\CT$, which we will refer to as $\CS$ (Patterns and Singletons). In Algorithm~\ref{alg:cov} we specify our \textsc{Cover} algorithm, which iterates through $\CS$ in a pre-specified order that we will detail later. For each pattern $X$ the algorithm greedily covers all occurrences that do not overlap with already covered data, and have fewer than $t(X)$ gaps. To bound the number and size of gaps we only allow occurrences where the number of gap positions is smaller than the length of the pattern itself. This avoids gap codes becoming cheaper than fill codes, by which occurrences with more gaps would be preferred over compact occurrences. This process continues until all data is covered. In Figure~\ref{fig:toy1} we show in 3 steps how a $\CS$ is used to cover the example dataset. 
			
			\vspace{0.5em}\noindent\textit{Example 1.} In Algorithm~\ref{alg:cov}, we start with an empty cover, corresponding to an uncovered dataset as in the top left of Fig~\ref{fig:toy1}. Next, we cover the data using the first pattern from the $\CS$ (line 1). For each occurrence of the pattern \{$\mev{a}{d}$\} (line 2), we add it to our cover (line 4) when it does not overlap previously added elements and has a limited number of gaps (line 3). In Step 1 of Fig~\ref{fig:toy1} we show the result of covering the data with the pattern \{$\mev{a}{d}$\}. We continue to do the same for the next pattern \{$h,e$\} in $\CT$. This gives us the cover as shown in Step 2. With only the singleton patterns left to consider in $\CT$ there is only one way to complete our cover by using them for all so far uncovered events. Now all events in the dataset are covered, thus we can break out of our loop (line 6) and return the final cover of Step 3 in Fig.~\ref{fig:toy1} (line 7).
			
			\begin{algorithm}[t]
				\caption{The \textsc{Cover} Algorithm}
				\label{alg:cov}
				\begin{algorithmic}[1]
					\INPUT A sequence $\S$, a set of patterns $\CS$, and $\C=\emptyset$
					\OUTPUT A cover $\C$ of $\S \in \DB$ using $\CS$
					\For{each $X \in \CS$ in order}		\Comment{see Sec.~\ref{sec:alg} for order}
						\ForAll{$o \in \occs(X,S)$} 	\Comment{all occurrences of $X$}						
							\If{$\C \cap o = \emptyset$ { \textbf{and} } $t(o) < 2 t(X)$ }
								\State $\C \gets \C \cup o$		
							\EndIf
						\EndFor
						\If{$\C$ completely covers $\S$}
							\State \textbf{break}
						\EndIf
					\EndFor
					\State {\textbf{return}} $\C$
				\end{algorithmic}
			\end{algorithm}

		\subsubsection{Encoding} \label{subsec:enc}	
			A cover $\C$ specifies how the data will be encoded. Conceptually, we can decompose it into a pattern code stream and a gap code stream. The pattern stream $\Cp$ contains a sequence of pattern codes, $\codepattern(X)$ for pattern $X \in \CT$, used to describe the data, whereas the gap code stream $\Cg$ consists of codes $\codegap(X)$ and $\codefill(X)$ indicating whether and where a pattern instance contains gaps or not.
			
			To encode a dataset we traverse our dataset from left-to-right and top-to-bottom and each time we encounter a new pattern in our cover we add the code for this pattern to $\Cp$. When moving to the next multi-event in the dataset we add for each pattern, that we already encountered but not yet completely passed by, a gap or fill code to $\Cg$. We choose a gap code for a pattern if the current multi-event does not contain any event from the pattern or a fill code if it does. We are finished when we have encoded all multi-events in the data. 
		
\begin{figure}[h]
				\begin{center}			
					\includegraphics[width=0.48\textwidth]{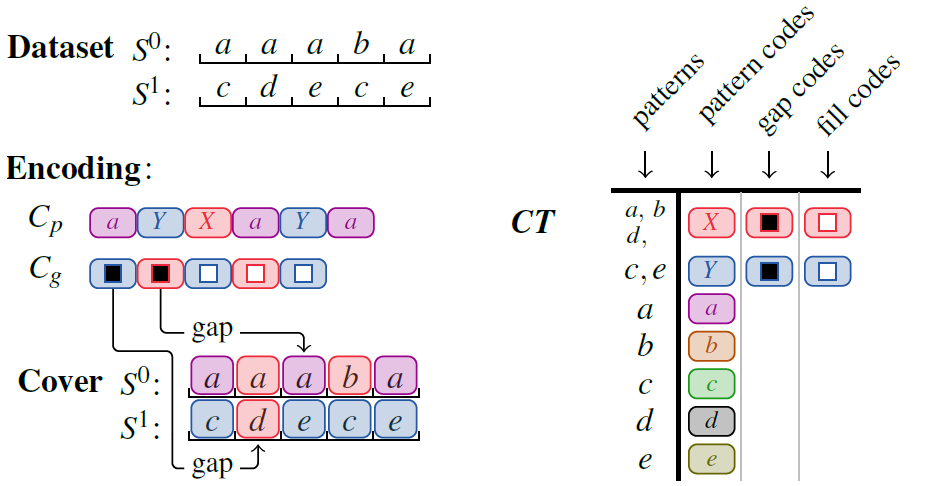}
				\end{center}
				\caption{The encoding of the dataset given a cover and $\CT$. See Example~2 for more details.}
				\label{fig:toy2}
			\end{figure}

			\vspace{0.5em}\noindent\textit{Example 2.} In Figure~\ref{fig:toy2} we show how a cover is translated to an encoding. To encode the first multi-event we first add $\code_p(a)$ and then $\code_p(Y)$ to $\Cp$. For the second multi-event we add $\code_p(X)$ to $\Cp$ and $\codegap(Y)$ to $\Cg$, because event $e$ from pattern $Y$ does not yet occur in this multi-event. For the third multi-event we add $\code_p(a)$ to $\Cp$, and $\codegap(X)$ and $\codefill(Y)$ to $\Cg$. Code $\codegap(X)$ marks the gap for pattern $X$ and $\codefill(Y)$ indicates that event $e$ from pattern $Y$ does occur in this multi-event. The fourth multi-event results in the addition of $\code_p(Y)$ to $\Cp$ and $\codefill(X)$ to $\Cg$, where the latter marks the presence of event $b$ from pattern $X$ in this multi-event. For the last multi-event we add $\code_p(a)$ to $\Cp$ and $\codefill(Y)$ to $\Cg$, which completes the encoding.
		
			Using $\Cp$ and $\Cg$ we can compute the actual codes to construct the code table $\CT$ corresponding to $\CS$ used to cover the data. We will encode the data using optimal prefix codes~\cite{cover:06:elements}, the length of which we can compute by Shannon entropy. In our case this means that the length of an optimal pattern code for pattern $X$ is the negative log-likelihood of a pattern in the cover \cite{grunwald:07:book}, that is
			\[
				L(\codepattern(X \mid \CT)) = - \log \bigg( \frac{\usage(X)}{\sum_{Y \in \CT} \usage(Y)} \bigg)  \quad ,
			\]
			where $\usage(X)$ is the number of times a pattern $X$ is used to describe the data, i.e.\ $\usage(X) = |\{Y \in \Cp | Y = \codepattern(X)\}|$.
					
			Gap and fill code lengths are computed similarly, corresponding to the negative log-likelihood of these codes within the usages of the corresponding pattern. That is, we have
			\[
				L(\codegap(X \mid \CT)) = - \log \bigg( \frac{\gaps(X)}{\gaps(X) + \fills(X)} \bigg) \quad , 
			\]
			\[
				L(\codefill(X \mid \CT)) = - \log \bigg( \frac{\fills(X)}{\gaps(X) + \fills(X)} \bigg) \quad .
			\]
			where $\gaps(X)$ and $\fills(X)$ are the number of times a gap, resp.\ fill-code of pattern $X$ is used in the cover of the data, i.e.\ $\gaps(X) = |\{Y \in \Cg \mid Y = \codegap(X)\}|$ and analogue for $\fills(X)$.

		\subsubsection{Encoded Length of Data Given Code Table} \label{sec:enc-db}
			Now that we have determined the cover and encoding scheme, we can formalise the calculation of the encoded length of the data given the $CT$. This encoded length is the sum of the encoded length of following terms: the pattern stream, the gap stream, the number of attributes, the number of sequences and the length of each sequence. Formally, we have
			\begin{align}
				L(D \mid \CT) =\ & L(\Cp \mid \CT) + L(\Cg \mid \CT) + L_{\mathbb{N}}(|A|)\\
							  &+ \LN(|D|) + \sum_{S\in \DB} \LN(|S|)
			\quad ,
			\end{align}
			where $\LN$ is the MDL optimal Universal code for integers~\cite{grunwald:07:book} and the encoded length of the pattern and gap stream are simply the code lengths of all codes in the streams, 		
			\begin{align} 
				L(\Cp \mid \CT) =\ \sum_{X \in \CT} & \usage(X)  L(\codepattern(X))\\ 
				L(\Cg \mid \CT) =\ \sum_{X \in \CT}^{t(X)>1} & \big[ \gaps(X) L(\codegap(X))\\
								&\ + \fills(X) L(\codefill(X)) \big] \quad .
			\end{align}
			
		\subsubsection{Encoded Length of Code Table} \label{sec:enc-ct}
			The encoded length of the code table consists of the following parts. For each attribute $A_j$ we encode the number of singletons and their support in $\DB^j$. Then we encode the number of non-singleton patterns in the code table, the sum of their usages, and then using a strong number composition their individual usages. Last, we encode the patterns themselves using $L(X \in \CT)$.  We hence have
			\begin{align}
				L(\CT \mid \C) 	=\ & \sum_{j \in |A|}{\bigg( \LN(|\Omega_j|) + \log \binom{|\DB^j|}{|\Omega_j|}\bigg)}\\			
								&\ + \LN(|\P| + 1) + \LN(\usage(\P) + 1)\\
								&\ + \log \binom{\usage(\P)}{|\P|} + \sum_{X \in \P} L(X \in \CT)  \quad .
			\end{align}
			For the encoded length of a non-singleton pattern $X \in \CT$ we have
			\begin{align}
				L(X \in \CT) =\ & \LN(t(X))\ + \LN(\gaps(X) + 1)\\
				&\ +\ \sum_{t(X)} \log(|A|) + \sum_{x \in X} L(\codepattern(x \mid \ST)) \quad ,
			\end{align}
			where we first encode its length, and its total number of gaps -- note that we can derive the total number of fills for this pattern from these two values. As $\LN$ is defined for integers $z \geq 1$, we apply a $+1$ shift wherever $z$ can be zero~\cite{grunwald:07:book}.
			Then, per time step, we encode for how many attributes the pattern defines a value, and what these values are using the singleton-only, or Standard Code Table ($\ST$). For the encoded length of an event given the $\ST$ we have
			\begin{align}
				L(\codepattern(x \mid \ST)) = -\log \bigg( \frac{\support(x \mid D)}{||D||} \bigg)  \quad ,
			\end{align}
			which is simply the negative log-likelihood of the event under an independence assumption.

	\subsection{Formal Problem Definition}
		Loosely speaking, our goal is to find the most succinct description of the data. By MDL, we define the optimal set of patterns for the given data as the set for which the optimal cover and associated optimal code table minimise the total encoded length. As such we have the following problem definition.

\vspace{0.5em}
		\noindent\textbf{Minimal Pattern Set Problem} \emph{Let $\Omega$ be a set of events and let $\DB$ be a multivariate sequential dataset over $\Omega$, find the minimal set of multivariate sequential patterns $\P$ and cover $\C$ of $\DB$ using $\P$ and $\Omega$, such that the encoded length $L(\CT, \DB) = L(\CT \mid \C) + L(\DB \mid \CT)$ is minimal, where $\CT$ is the code-optimal code table for $\C$.}

	\vspace{0.5em}
	Let us consider how complex this problem is. 	
		Firstly, the number of possible pattern sets (with a maximum pattern length of $n$) is
		\[
			\sum_{k=1}^{2^{|\Omega|^{n}} - |\Omega| - 1} \binom{2^{|\Omega|^{n}} - |\Omega| - 1}{k} \quad .
		\]
		Secondly, to use a pattern set to cover the data we also need to specify the order of the patterns in the set. That is, we need to find the optimal order for the elements in the pattern set to find the one that minimises the total encoded length. The total number of ways to cover the dataset using one of the possible ordered pattern sets is
		\[
			\sum_{k=1}^{2^{|\Omega|^{n}} - |\Omega| - 1} \binom{2^{|\Omega|^{n}} - |\Omega| - 1}{k} \times (k + |\Omega|)!  \quad .
		\]
			
		Moreover, unfortunately, it does not show submodular structure nor (weak) (anti-)monotonicity properties by which we would be able to prune large parts of it. Hence, we resort to heuristics.

\section{The Ditto Algorithm} \label{sec:alg}	
	
In this section we present \ourmethod, an efficient algorithm to heuristically approximate the MDL optimal summary of the data.
	In particular, it avoids enumerating all multivariate patterns, let alone all possible subsets of those. Instead, it considers a small but highly promising part of this search space by iterative bottom-up search for those patterns that maximally improve the description. More specifically, we build on the idea of \textsc{Slim}~\cite{smets:12:slim} and \textsc{sqs}~\cite{tatti:12:sqs}. That is, as candidates to add to our model, we only consider the most promising combinations of already chosen patterns -- as identified by their estimated gain in compression. 

	We give the pseudo code of \ourmethod as Algorithm~\ref{alg:main}. We start with singleton code table $ST$ (line~\ref{ln:main:st}) and a set of candidate patterns of all pairwise combinations of singletons  (line~\ref{ln:main:cand}). We then iteratively add patterns from the candidate set to code table $\CT$ in $\textbf{Candidate Order}$: $\downarrow \mathit{estimated~gain}(X)$, $\downarrow \support(X \mid \DB)$, $\downarrow ||X||$, $\downarrow L(X \mid \ST)$ and $\uparrow$ lexicographically (line~\ref{ln:main:co}). This order prefers the most promising candidates in terms of compression gain. When a new pattern improves the total encoded length $L(\DB, \CT)$ we keep it, otherwise we discard it (line~\ref{ln:main:l}). After acceptance of a new pattern we prune (line~\ref{ln:main:pr}) $\CT$ and recursively test whether to add variations (line~\ref{ln:main:var}) of the accepted pattern in combination with its gap events. When all variations are tested recursively, we update the candidate set by combining $\CT \times \CT$ (line~\ref{ln:main:up}).
	
	We give the details of each of these steps below, as well as explain how to gain efficiency through smart caching, and discuss the computational complexity of \ourmethod.
		
	\begin{algorithm}[t]
		\caption{The \ourmethod Algorithm}
		\label{alg:main}
		\begin{algorithmic}[1]
			\INPUT The dataset $\DB$ and singleton code table $\ST$
			\OUTPUT An approximation to the \textbf{Minimal Pattern Set Problem} 
			\State $\CT \gets \ST$											\label{ln:main:st}
			\State $\cand \gets \CT \times \CT$								\label{ln:main:cand}
			\For{$X \in \cand$ in \textbf{Candidate Order}}					\label{ln:main:co}
				\If{$L(\DB, \CT \oplus X) < L(\DB, \CT)$}							\label{ln:main:l}
					\State $\CT \gets \textsc{Prune}(\DB, \CT \oplus X)$			\label{ln:main:pr}
					\State $\CT \gets \textsc{Variations}(\DB, X, \CT)$		\label{ln:main:var}
					\State $\cand \gets \CT \times \CT$						\label{ln:main:up}
				\EndIf	
			\EndFor
			\State {\bf return} $\CT$
		\end{algorithmic}
	\end{algorithm}
	
	\subsection{Covering}
	As detailed in the previous sections, to compute $L(\DB, \CT)$, we first need to cover the data with $\CT$. The \textsc{Cover} algorithm covers the data using the patterns as they are ordered in $\CT$. To find the optimal cover we need to identify the cover order that leads to the smallest encoded length of the data. We do so greedily, by considering the pattern set in a fixed order. As our goal is to compress the data, we prefer patterns that cover many events with just a short code length. We hence define the \textbf{Cover Order} as follows: $\downarrow ||X||$, $\downarrow \support(X \mid \DB)$, $\downarrow L(X \mid \ST)$ and $\uparrow$ lexicographically. It follows the intuition that we give preference to larger and more frequent patterns, for which we expect a higher compression gain, to cover the data first, as these maximise the likelihood of the cover. 

	\subsection{Candidates and Estimation}
\begin{figure}[t]
			\begin{center}
				\includegraphics[width=0.48\textwidth]{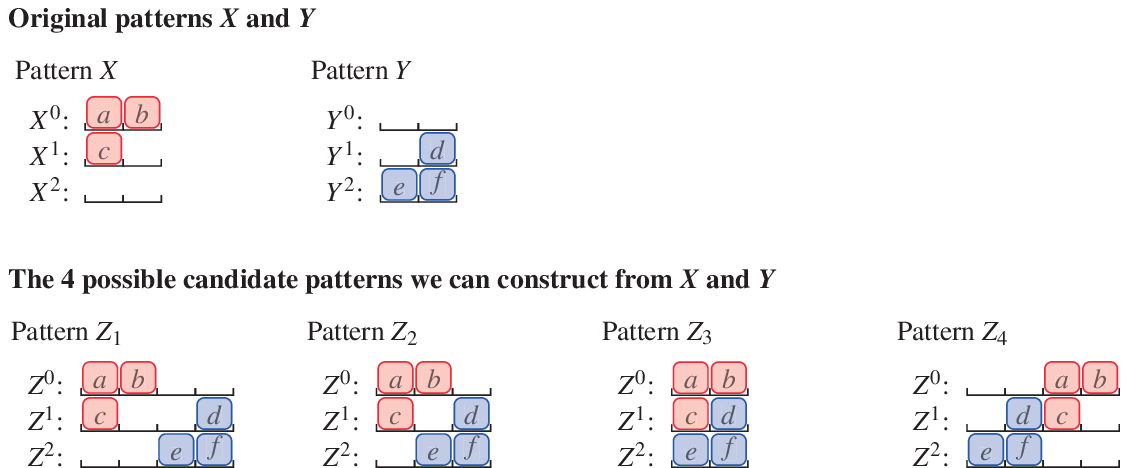}
			\end{center}
			\caption{The 4 possible candidate patterns constructed from different alignments of $X$ and $Y$.}
			\label{fig:align}
		\end{figure}

		Conceptually, at every iteration the set of candidates from which we can choose, consists of the Cartesian product of the code table with itself. Two patterns $X$ and $Y$ can be combined to form new candidate patterns. Each alignment of $X$ and $Y$ without gaps in either $X$, $Y$ and the resulting pattern, forms a new candidate, with the exception of alignments in which $X$ and $Y$ overlap. See Figure~\ref{fig:align} for an example.
		
		Selecting the candidate with the highest gain is very expensive -- we would need to cover the whole data for every candidate. Instead, we select the candidate with the highest \emph{estimated} gain -- which can be done much more efficiently. Intuitively, based on the estimated gain ($\Delta L'$) we only consider candidate patterns in a lazy fashion based on their usage and do not consider patterns with a lower usage than the current best candidate. For notational brevity, for a pattern $X \in \CT$ we use $x = \usage(X)$. Further, let $s$ be the total usage count of all patterns in $\CT$, i.e.\ $s = \sum_{X \in \CT} x$. For $\CT' = \CT \oplus Z$, we use $x'$ and $s'$ similarly. We estimate $z$, the usage of $Z$,  optimistically as the minimum of the usages of $X$ and $Y$ -- and as $x/2$ when  $X = Y$ (because the usage of $XX$ can not be higher). Formally, our gain estimate $\Delta L'(\CT \oplus Z,D)$ of the true gain $\Delta L(\CT \oplus Z,D)$ for adding pattern $Z = X \cup Y$ to $\CT$ is as follows,
		\begin{align}
		\Delta L'(\CT',\ \DB) 	=\ &\ \Delta L'(\CT' \mid \DB) + \Delta L'(\DB \mid \CT') \quad ,\\
		\Delta L'(\CT' \mid \DB)=\ &- \LN(|Z|) - \sum_{l(Z)} \log(|A|)\\
								   &- \sum_{Z[i] \in Z}\sum_{z \in Z[i]} L(\codepattern(z \mid \ST)) \quad , \\
		\Delta L'(\DB \mid \CT')=\ &\ s \log s - s' \log s' + z \log z - x \log x\\
								   &+ x' \log x' - y \log y + y' \log y' \quad .
		\end{align}
		That is, the estimated gain of adding pattern $Z$ to $\CT$ thus consists of the estimated gain in the size of the data, minus the increase in the size of $\CT$. Note that, $\Delta L'$ is an estimate and for simplicity we ignore the effects of adding pattern $Z$ to code table $\CT$ on the pattern and (no-)gap usages of patterns other than $X$ and $Y$. 

	\subsection{Pruning}
		After the acceptance of a new pattern in our code table other patterns may have become redundant as their role may have been overtaken by the newer pattern. Therefore, each time a pattern $X$ is successfully added to the code table, we consider removing those $Y \in \CT$ for which the usage decreased and hence the pattern code length increased. Algorithm~\ref{alg:pru} describes how a code table is pruned. 
		
		\begin{algorithm}
			\caption{The \textsc{Prune} Algorithm}
			\label{alg:pru}
			\begin{algorithmic}[1]
				\INPUT The dataset $\DB$ and a code table $\CT$
				\OUTPUT A pruned code table
				\State $\cand \gets X \in \CT$ with decreased usage						\label{ln:pru:cand}
				\For{$X \in \cand$ in \textbf{Prune Order}}								\label{ln:pru:po}
					\If{$L(\DB, \CT \setminus X) < L(\DB, \CT)$}							\label{ln:pru:l}
						\State $\CT \gets \CT \setminus X$								\label{ln:pru:del}
						\State $\cand \gets \cand \cup \{ Y \in \CT \mid $ usage decreased$\}$	\label{ln:pru:up}
					\EndIf
					\State $\cand \gets \cand \setminus X$								\label{ln:pru:disc}
				\EndFor
				\State {\bf return} $\CT$
			\end{algorithmic}
		\end{algorithm}
		
	\subsection{Generating Pattern Variations}
		To efficiently discover a large and diverse set of promising patterns -- without  breadth-first-search, which takes long to find large patterns, and without  depth-first-search, which would be prohibitively costly -- we consider \emph{variations} of each accepted pattern to the code table. That is, when a pattern leads to a gain in compression, we consider all ways by which we can extend it using events that occur in the gaps of its usages. This way we consider a rich set of candidates, plus speed up the search as we are automatically directed to patterns that actually exist in the data. Algorithm~\ref{alg:var} outputs a code table possibly containing variations of the lastly added pattern $Y$. 
	
		\begin{algorithm}
			\caption{The \textsc{Variations} Algorithm}
			\label{alg:var}
			\begin{algorithmic}[1]
				\INPUT The dataset $\DB$, a pattern $Y$ and a code table $\CT$
				\OUTPUT A code table possibly containing variations of $Y$
				\State $\cand \gets Y \times \mathit{gap~events}(Y)$					\label{ln:var:cand}
				\For{$X \in \cand$}												\label{ln:var:loop}
					\If{$L(\DB, \CT \oplus X) < L(\DB, \CT)$}							\label{ln:var:l}
						\State $\CT \gets \textsc{Prune}(\DB, \CT \oplus X)$			\label{ln:var:pr}		
						\State $\CT \gets \textsc{Variations}(\DB, X, \CT)$		\label{ln:var:var}
					\EndIf
					\State $\cand \gets \cand \setminus X$
				\EndFor
				\State {\bf return} $\CT$
			\end{algorithmic}
		\end{algorithm}
		
		For example, consider the dataset \{$a, b, a, b, c, a, c, a$\} where pattern \{$a, a$\} occurs twice with a gap of length one. After adding pattern \{$a, a$\} to $\CT$ we consider the patterns \{$a, b, a$\} and \{$a, c, a$\} for addition to $\CT$.
	\newpage	
	\subsection{Faster Search through Caching}		
		The space of all possible multivariate patterns is extremely rich. Moreover, in practice many candidate patterns will not lead to any gain in compression. In particular, those that occur only very infrequently in the data are unlikely to provide us any (or much) gain in compression. We thus can increase the efficiency of \ourmethod by allowing the user to impose a minimum support threshold for candidates. That is, only patterns $X$ will be evaluated if they occur at least $\sigma$ times in the data. To avoid time and again re-evaluating candidates of which we already know that they are infrequent, we cache these in a tree-based data structure. Only \emph{materialised} infrequent pattern are added to the tree. Future candidates are only materialised when none of its subsets are present in the tree, as by the a priori principle we know it can not be frequent~\cite{mannila:94:efficient}.
		
		Even though this tree can theoretically grow very large, in practise it stays relatively small because we only consider a small part of the candidate space. That is, we only combine patterns we know to be frequent to form new candidate patterns. In practice, we found that \ourmethod only has to cache up to a few thousand candidates. Using this tree we see speed ups in computation of 2 to 4 times, while also memory consumption is strongly reduced. For some datasets the difference is even bigger, up to an order of magnitude.

		In this work we only consider keeping track of infrequent candidates. Note, however, that at the expense of some optimality in the search additional efficiency can be gained by also storing \emph{rejected} candidates in the tree. In both theory and practice, however, candidates rejected in one iteration may lead to compression gain later in the process~\cite{smets:12:slim}.

	\subsection{Complexity}		
		The time complexity of \ourmethod has a simple upper bound as follows. In the worst-case we cover the data for each frequent pattern from the set of all frequent patterns $\mathcal{F}$ in each iteration. Covering takes $O(|\CT| \times ||\DB||)$ and the number of iterations is worst-case $O(\mathcal{|F|}$). Together, the worst-case time complexity is
		\[O( |\mathcal{F}|^2 \times |\CT| \times ||\DB||) \quad .\]
		From the experiments in Section~\ref{sec:exp}, however, we will learn that this estimate is rather pessimistic. In practice the code table stays small ($|\CT| \ll |\mathcal{F}|$), we only consider a subset of all frequent patterns and we do this greedily. In practice the runtime of \ourmethod therefore stays in the order of seconds to minutes.

\section{Related Work} \label{sec:rel}

		The first to use MDL to summarise transaction data were Siebes et al.\ \cite{siebes:06:item}, resulting in \textsc{Krimp} \cite{vreeken:11:krimp}. It shifted focus from the long-standing goal of mining collections of patterns that describe the set of all frequent patterns, e.g.\ closed \cite{pasquier:99:discovering} frequent patterns, to a \emph{pattern set} that describes the data. 
		
		Similar to the transaction data domain, summarisation of sequential data also developed from frequent pattern mining. For sequential data, the key types of patterns studied are frequent subsequences~\cite{pei:04:pspan}, and frequent episodes~\cite{agrawal:95:mining,mannila:97:discovery}. As with all traditional pattern mining approaches, redundancy is also a key problem when mining sequential patterns~\cite{wang:04:bide,he:11:dasfaa}. To this end, Tatti and Vreeken \cite{tatti:12:sqs} proposed to instead approximate the MDL-optimal summarisation of event sequence data using serial episodes. Their method \textsc{Sqs} deals with many challenges inherent to this type of data, such as the importance of the order of events and the possibility for patterns to allow gaps in their occurrences -- aspects we build upon and extend. Other methods exist, but either do not consider \cite{bathoorn:06:freqpatset,mannila:00:global} or do not punish gaps \cite{lam:12:gokrimp,lam:14:seq} with optimal codes. None of these methods consider, or are easily extended to multivariate sequential data. 

		One of the first to consider multivariate sequential patterns, by searching for multi-stream dependencies, were Oates et al.\ \cite{oates:96:structure}. Tatti and Cule~\cite{tatti:11:multievent} formalised how to mine the set of closed frequent patterns from multivariate sequential data, where patterns allow simultaneous events. In \cite{wu:13:utility} the mining of high utility sequential patterns is studied, where they allow simultaneous events. Chen et al.\ \cite{chen:10} and Moerchen et al.\ \cite{morchen:07} study mining interval-based patterns, where frequency is determined by how often univariate patterns co-occur within a given interval. All these methods are traditional pattern mining techniques in the sense that they return all patterns that pass an interestingness threshold. 

		Whereas traditional pattern mining techniques often only consider discrete data, there does exist extensive literature on mining patterns in continuous valued time series. These patterns are often called `motifs' \cite{lin:02:motifs,chiu:03:motifs}. For computational reasons, many of these methods first discretise the data~\cite{lin:07:sax}. Most motif discovery algorithms consider only univariate data. Example proposals for motif discovery in a multivariate setting include that by Tanaka et al.\ \cite{tanaka:05:mdlmotif}, who first transform the data into one dimension before the pattern discovery process and do not consider gaps, and by \cite{minnen:07:detecting}, who do not allow patterns to temporally overlap even if they span different dimensions and do not consider variable-length motifs. More recently  Vespier et al.\ \cite{vespier:13:mcm}, mine characteristic multi-scale motifs in sensor-based time series but aim at mining all motifs, not a succinct summary.

		To the best of our knowledge there are no methods yet to summarise \emph{multivariate} sequential data, other than regarding each attribute separately or with restrictions on the pattern language \cite{bertens:14}. In this work we introduce \ourmethod to discover important sequential associations between attributes by mining succinct summaries using rich multivariate patterns.

\section{Experiments} \label{sec:exp}
	
		We implemented \ourmethod in C++ and generated our synthetic data and patterns using Python. We make our code available for research purposes.\!\footnote{\oururl} All experiments were conducted on a 2.6 GHz system with 64 GB of memory. 
		For our experiments on real data we always set the minimum support threshold as low as feasible, unless domain knowledge suggests otherwise.

		We evaluate \ourmethod on a wide range of synthetic and real world data. As discussed in Sec.~\ref{sec:rel}, there exist no direct competitors to \ourmethod. Traditional pattern mining and motif discovery methods `simply' mine all patterns satisfying some constraints. For summarising sequential data, most existing methods consider univariate data~\cite{tatti:12:sqs,lam:12:gokrimp}. The only summarisation approach for multivariate sequential data considers the special case where attributes are ordered (e.g.\ frequency bands)~\cite{bertens:14}, whereas we consider multivariate sequential data in general. We empirically compare \ourmethod to \textsc{Sqs}~\cite{tatti:12:sqs}. We do so by applying \textsc{Sqs} to each univariate sequence $S^j \in \DB$, combining these results into one pattern set.
		
	\subsection{Synthetic Data}
		To validate whether \ourmethod correctly identifies true multivariate sequential patterns from a dataset, we first consider synthetic data. In particular, we generate random data in which we plant a number of randomly generated patterns of different characteristics. Clearly, ideally the true patterns are recovered. Moreover, ideally no other, spurious patterns that are only due to noise are returned. To this end we perform an extensive set of experiments varying the number of events, the number of attributes and the alphabet size of the dataset, and the number, frequency and size of the planted patterns. 
		
\subsubsection*{Data Generation}
As noted in Table~\ref{tab:syn}, for each experiment we generated $t(\DB)$ random multi-events on $|A|$ attributes (i.e.\ a total of $||D||$ events) with an alphabet size per attribute of $|\Omega_i|$. Further, after the data generation $|\P|$ patterns are planted, where each pattern $X$ has a size $||X||$, a 5\% chance on a gap between subsequent multi-events, and a support such that each pattern spans $\support$\% of all events in the dataset. An example of an insertion of a pattern in a random dataset that does not lead to an actual occurrence of that pattern is when due to the gap chance the minimal window of the pattern contains too many gaps. We do not allow patterns to overwrite each other during generation, as this makes evaluation much more complicated -- i.e.\ it becomes unclear whether not recovering a pattern is an artefact of the search or of the data generation process. Further, only for experiments with 50 attributes, we prevented that pattern occurrences interleave and did not allow an event to be used in more than one pattern to assure that the planted patterns are actually present in the data. This restriction is justified because we are merely testing whether our algorithm is able to find multivariate patterns in multivariate data and without it patterns will easily crowd each other because of the high number of attributes.

\subsubsection*{Evaluation}
We evaluate the quality of the pattern set discovered by considering how close they represent the planted patterns. In particular, following~\cite{webb:14:selfsufs} we consider both exact ($=$) and subset ($\subset$) matches. Exact indicates that the pattern exactly corresponds with a planted pattern, whereas subset implies that it is only part of a planted pattern. 
In addition, we consider how well the planted patterns are recovered; we report how many of the events of the ground truth pattern set $\P$ we can cover with the discovered patterns. The higher this ratio, the better this result. Last, we consider the gain in compression of the discovered model over the initial, Standard Code Table. The higher this number, the better -- the best score is attained when we recover all patterns exactly, and no further noise. 
		
\subsubsection*{Results}

We first consider the traditional approach of mining all (closed) frequent multivariate patterns. We do so using the implementation of Tatti and Cule~\cite{tatti:11:multievent}. We use a minimal support of 90\% of the lower-support planted pattern. This choice is made to ensure that even when not all insertions of a pattern result in an actual occurrences, it can still be discovered. For the most simple synthetic dataset we consider, corresponding to the first row of Table~\ref{tab:syn}, this takes a few days, finally reporting a \emph{lower bound} of 14\,092\,944\,394 frequent patterns, and returning 6\,865 closed frequent patterns -- hardly a summary, knowing there are only 5 true patterns. In the remainder we therefore do not consider traditional pattern mining. 

Next, we consider \ourmethod and \Sqs. We report the results in Table~\ref{tab:syn}. On the right hand side of the table we see that \ourmethod recovers all planted patterns, and does not report a single spurious pattern (!). In all cases it recovers the ground truth model, and obtains the best possible gain in compression. Next to the exactly identified planted pattern sometimes it also identifies some subsets of the planted patterns. This is a result of the data generation, i.e. subsets are sometimes included in the code table when planted occurrences contain too many gaps to be covered with the exact pattern. 
The patterns \Sqs discovers, on the other hand, are only small univariate fragments of the ground truth, recovering roughly only 10\% to 30\% of the ground truth. The near-zero gains in compression corroborate it is not able to detect much structure. 
		
Regarding runtime and scalability, \ourmethod scales very favourably. Although \Sqs is faster it considers only the much simpler case of univariate data and patterns. \ourmethod requires seconds, up to a few minutes for the largest data, even for very low minimal support thresholds. Analysing the runtime of \ourmethod in more detail show how well its heuristics work; most time is spent on computing the minimal windows for candidates, of which up to ten thousand are materialised. Only for a few hundred of these a full cover of the data is required to evaluate their contribution to the model. Smart implementation for computing the minimal windows of all candidates in one pass will hence likely speed up \ourmethod tremendously.

		\begin{table*}[h] 	
			\centering
			\caption{\ourmethod discovers all planted patterns on all synthetic datasets, without picking up on noise. Given are the base statistics of the synthetic datasets, and the results of \Sqs and \ourmethod.
			For \Sqs and \ourmethod we give the number of exactly recovered patterns ($=$) and the number of discovered patterns that are subsets of planted patterns ($\subset$). Further, we report how much of the ground truth is recovered ($R\%$), as well as the gain in compression over the singleton-only model ($L\%$), for both higher is better. Last, we give the runtime in seconds.}
			\label{tab:syn}				
	\begin{tabular}{rrrr rrr rrrrr rrrrr} 
	\toprule
	\multicolumn{4}{l}{\textbf{Data}} &
	\multicolumn{3}{l}{\textbf{Planted Patterns}} &
	\multicolumn{5}{l}{\textbf{\Sqs}} &
	\multicolumn{5}{l}{\textbf{\ourmethod}} \\
	\cmidrule(lr){8-12}
	\cmidrule(lr){13-17}
	\cmidrule(r){1-4}
	\cmidrule(lr){5-7}
	$||D||$ & $t(\DB)$ & $|A|$ & $|\Omega_i|$ &
	$|\P|$ & $||X||$ & $\support$ &
	$=$ & $\subset$ & $R\%$ & $\Delta L\%$ & time &
	$=$ & $\subset$ & $R\%$ & $\Delta L\%$ & time \\
	\midrule
	
		$100\,000$ 		& $10\,000$ 			& $10$ 			& $100$ 			& $5$ 			& 3--7 		& $1\%$ 			& 0 & 1 & 9.5 	& 0.1 	& 3 	& 5 & 0 & 100.0 & 3.0 	& 2 \\
		$100\,000$ 		& $10\,000$ 			& $10$ 			& $100$ 			& $5$ 			& \textbf{5}& $\mathbf{10\%}$ 	& 0 & 0 & 0 	& 0 	& 3 	& 5 & 5 & 100.0 & 31.6 	& 31 \\
		$100\,000$ 		& $10\,000$ 			& $10$ 			& $\mathbf{1\,000}$ & $5$ 			& 3--7 		& $1\%$ 			& 0 & 1 & 8.0 	& 0 	& 12 	& 5 & 0 & 100.0 & 2.9 	& 4 \\
		$100\,000$ 		& $10\,000$ 			& $10$ 			& $100$ 			& $\mathbf{20}$ & 3--7 		& $1\%$ 			& 0 & 9 & 19.5 	& 0.9 	& 4 	& 20& 0 & 100.0 & 12.9  & 15\\
		$500\,000$ 		& $10\,000$ 			& $\mathbf{50}$ & $100$ 			& $5$ 			& 3--7 		& $1\%$ 			& 0 & 0 & 0 	& 0 	& 15 	& 5 & 1 & 100.0 & 3.1 	& 41\\
		$500\,000$ 		& $\mathbf{50\,000}$ 	& $10$ 			& $100$ 			& $5$ 			& 3--7 		& $1\%$ 			& 0 & 1 & 10.0 	& 0.2 	& 4 	& 5 & 0 & 100.0 & 3.1 	& 28 \\
		$1\,000\,000$ 	& $\mathbf{100\,000}$ 	& $10$ 			& $100$ 			& $5$ 			& 3--7 		& $1\%$ 			& 0 & 4 & 31.8 	& 0.3 	& 4 	& 5 & 14& 100.0 & 3.3 	& 374 \\
		
			\bottomrule		& & & & & & & 
\end{tabular}
		\end{table*}

	\subsection{Real World Data}
		As case studies we consider 4 datasets. An ECG sensor dataset, a structural integrity sensor dataset of a Dutch bridge, the text of the novel Moby Dick tagged with part-of-speech tags, and a multilingual text of an European Parliament resolution. See Table~\ref{tab:realprop} for their characteristics. 

		\begin{table*}[h]
			\centering
			\caption{Results on 4 real datasets. We give the basic statistics, as well as the number of patterns and relative gain in compression for both \Sqs and \ourmethod. The high compression gains show that \ourmethod discovers much more relevant structure than \Sqs.}

			\begin{tabular}{l rrrrr rrr rrr}					
				\toprule
				\multicolumn{6}{l}{\textbf{Data}} & \multicolumn{3}{l}{\textbf{\Sqs}} & \multicolumn{3}{l}{\textbf{\ourmethod}}\\
				\cmidrule(r){1-5}
				\cmidrule{7-9}
				\cmidrule(l){10-12}
				Dataset & $t(D)$ & $|D|$ & $|A|$ & $|\Omega|$ & $\support$ & $|\P|$ & $\Delta L\%$ & time ($s$) & $|\P|$  & $\Delta L\%$ & time ($s$) \\					
				\midrule					
				ECG			& $2\,999$	& 1		& 2	& 6			& 10	& 57 & 38.8 & 1 & 11 & \textbf{75.3} & 360 \\
				Bridge		& $5\,000$	& 1		& 2	& 10		& 100	& 21 & 58.8 & 1 & 22 & \textbf{76.3} & 325 \\		
				Moby Dick	& $2\,248$  & 103   & 2 & 887       & 5     & 20 & 1.7 	& 3 & 79 & \textbf{14.3}& 102 \\												
				Text		& $5\,960$	& 115	& 3	& $4\,250$ 	& 10	& 35 & 1.6 	& 12& 51 & \textbf{2.2} & 136 \\					
				\bottomrule & 
			\end{tabular}				
			\label{tab:realprop}
		\end{table*}

\begin{figure}[t]
				\centering
				\includegraphics[width=0.48\textwidth]{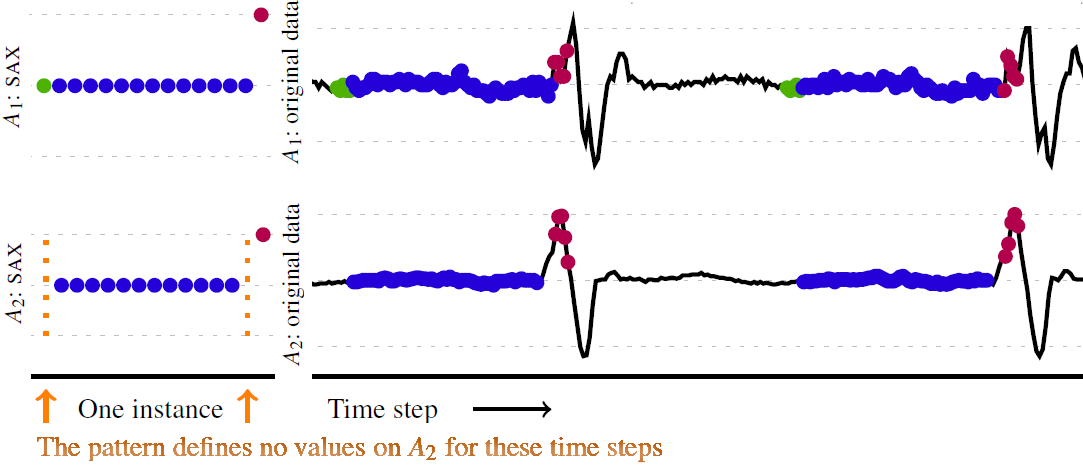}
				\caption{The top ordered pattern for the ECG data in the code table (left) and its first 2 occurrences in the data (right). The first time step of an occurrence is marked in green, subsequent ones in blue, and the last in  red. 
				}
				\label{fig:ecg-zoom}			
			\end{figure}

\paragraph{ECG}			
To investigate whether \ourmethod can find meaningful patterns on real-valued time series we consider a well-known ECG dataset.\!\footnote{ECG -- \url{physionet.org/physiobank/database/stdb}} 
			For ease of presentation, and as our main goal is to show that \ourmethod can discover multivariate patterns, we consider only two sensors.
			As preprocessing steps we applied 3 transformations: we subsampled the data, we transformed it from absolute to relative, and we discretised it using SAX~\cite{lin:07:sax}. 
			For the subsampling we replaced each 5 subsequent values with their average, thus creating a sequence 5 times shorter. Thereafter, we transformed the absolute data into relative data by replacing each value by the difference of its successor and its own value. Lastly, we discretised each attribute into 3 intervals using \Sax.
			Using a minimum support of 10, within 360 seconds \ourmethod discovers a code table containing 11 non-singleton patterns that capture the main structure of the data. In Figure~\ref{fig:ecg-zoom} we plotted 2 occurrences of the top ordered pattern of this code table. This is a multivariate pattern which shows a very characteristic repeating structure in the data comprising a longer flat line ending with a peak on both attributes simultaneously. Showing the power of our pattern language, note that the pattern does not define any values on the second attribute for the first and second-last time steps (indicated with arrows in Figure~\ref{fig:ecg-zoom}), while it does on the first attribute. This flexibility allows us to use this multivariate pattern to describe a larger part of the data.
			
			Like for the synthetic data, we also ran \Sqs. Again we see that as it cannot reward multivariate structure it does not obtain competitive gains in compression. Close inspection shows it returns many small patterns, identifying consecutive values.
			
\newpage
\paragraph{Bridge}
\label{subsec:infra}
			Next we consider the setting of monitoring the structural integrity of a bridge in the Netherlands.\!\footnote{Bridge -- \url{infrawatch.liacs.nl}} Amongst the collected sensor data are the strain, vibration and temperature. We selected 2 strain sensors (1 Hz) on consecutive pillars of the bridge and decomposed the signals into low, medium and high frequency components using techniques from \cite{vespier:12:timeseries}. We used the medium frequency components, after preprocessing, as our dataset. 
			As preprocessing, we transformed the absolute values into relative values by replacing each value by the difference of its successor and its own value. We then discretised each z-normalised attribute into 5 intervals using \Sax \cite{lin:07:sax}. 
			
			For a support threshold of 100, it takes \ourmethod 325 seconds to discover a code table with 22 non-singleton patterns. Although only one more than \Sqs at the same threshold, the patterns \ourmethod discovers are more descriptive. That is, they are multivariate and larger, leading to a much higher gain in compression. Moreover, the patterns it discovers correctly show the correlation between the attributes, whereas the patterns \Sqs discovers only identify univariate patterns.

\paragraph{Moby Dick}
			For more interpretable results, we next evaluated on text data. In particular we considered the first chapter of the book Moby Dick, written by Herman Melville,\!\footnote{\url{www.gutenberg.org}} aligning the text with part-of-speech tags.\!\footnote{http://nlp.stanford.edu/software/tagger.shtml}\footnote{https://gate.ac.uk/wiki/twitter-postagger.html} That is, one attribute comprises a stream of the words used in the book; each sentence is regarded as a sequence. The other attribute contains the tags that identify the type and function of each of these words. For example,
			\begin{center}
				\begin{tabular}{lccc} 					
					\textbf{attribute 1:} & VB   & PRP & NNP\\
					\textbf{attribute 2:} & Call & me  & Ishmael\\						
				\end{tabular}
			\end{center}
			for which we will further use the following notation, where each time step is enclosed by curly brackets and the symbols for different attributes within a time step are divided by a comma: \{VB, Call\}\{PRP, me\}\{NNP, Ishmael\}. A short description for the part-of-speech tags in this example can be found in Table~\ref{tab:pos}.
			
			With a support threshold of 5, it takes \ourmethod 102 seconds to discover 79 non-singleton patterns. After studying the resulting pattern set we found that the identified patterns show highly intuitive multivariate structure. The highest ordered patterns together with examples of text fragments that match these patterns are shown in Table~\ref{tab:md}.
			\begin{table}[h]
				\centering
				\caption{The highest ordered patterns in the code table for the Moby Dick dataset, together with example fragments of from the text which correspond to the patterns.}			
				\begin{tabular}{ll}
					\textbf{Pattern} & \textbf{Example text fragments} \\
					\{TO, to\}\{VB\}\{DT, a\}\{NN\} & to get a broom, to buy (him) a coat\\
					\{DT, the\}\{JJ\}\{NN\} 		& the green fields, the poor poet\\
					\{DT, a\}\{JJ\}\{NN\} 			& a mazy way, a hollow trunk  \\
					\{DT, the\}\{JJ\}\{NNS\} 		& the wild conceits, the old (sea-)captains\\
					\{PRP, i\}\{RB\}\{VB\} 			& I quietly take, I always go\\
					\{EX, there\}\{VBZ, is\} 		& there is\\
				\end{tabular}
				\label{tab:md}
			\end{table}

			\begin{table}[h]
				\centering
				\caption{A short description of the part-of-speech tags used in the examples in this paper for the Moby Dick experiment.
				}
				\begin{tabular}{ll} 
					\textbf{Tag}	& \textbf{Explanation} \\
					DT	& Determiner \\
					EX 	& Existential \emph{there} \\
					JJ	& Adjective\\
					NN	& Noun, singular or mass\\
					NNP	& Proper noun, singular\\
					NNS	& Noun, plural\\
					PRP	& Personal pronoun \\
					RB	& Adverb\\
					TO	& \emph{to} \\
					VB	& Verb, base form\\
					VBZ	& Verb, 3rd person singular present \\									
				\end{tabular}							
				\label{tab:pos}
			\end{table}
		
			Given the modest compression gain that \Sqs obtains, see Table~\ref{tab:realprop}, it is clear there is not much structure in each of the attributes separately; \ourmethod, however, is able to find a significant amount of multivariate structure.
			
\paragraph{Multilingual Text}
			As a final experiment, to further corroborate whether \ourmethod discovers meaningful and interpretable multivariate patterns, we consider mining patterns between the same text in different languages. To this end we collected a text in English, French and German from the European Parliament register of documents.\!\footnote{Text -- \url{www.europarl.europa.eu/RegistreWeb}} In this data we expect frequent combinations of words within one (or more) language(s), as well as (and of more interest), multivariate patterns in the form of translations between the languages. As a preprocessing step all text data was stemmed and stop words were removed. To keep the different languages aligned we regarded every paragraph as a subsequence and padded shorter aligned subsequences with sentinel events which are ignored by \ourmethod. This ensures that the difference in length of the sentences in different languages will not lead to very big misalignments. 
			
			For a support threshold of 10, \ourmethod takes 136 seconds to discover 51 non-singleton patterns. The highest ordered pattern, i.e.\ the one that aids compression most, is a translation pattern; it identifies the correct relation between the French word \emph{rel\`{e}ve}, the German phrase \emph{stellt fest dass} and the English word \emph{note}. Other high ordered patterns are the English \emph{EUR ($x$) million} and the German \emph{($x$) Millionen EUR}, and the words \emph{parliament}, \emph{Parlament} and \emph{parlement} in English, German and French, respectively. 

			The modest compression gain of \ourmethod over \Sqs, see Table~\ref{tab:realprop}, indicates this data is not very structured, neither univariately, nor multivariately. One of the reasons being the different order of words between different languages which results in very large gaps between translation patterns.

\section{Discussion} \label{sec:dis}
	Overall, the experiments show that \ourmethod works well in practice. In particular, the experiments on synthetic data show that \ourmethod accurately discovers planted patterns in random data for a wide variety of data and patterns dimensions. That is, \ourmethod discovers the planted patterns regardless of their support, their size, and the number of planted patterns -- without discovering any spurious patterns. \ourmethod also performed well on real data -- efficiently discovering characteristic multivariate patterns. 
	
	The results on the annotated text are particularly interesting; they give clear insight in non-trivial linguistic constructs, characterising the style of writing. Besides giving direct insight, these summaries have high downstream potential. One could, for example, use them for comparative analysis~\cite{budhathoki:15:diffnorm}. For example, for text identifying similarities and differences between authors, for sensor networks detecting and describing concept drift over time, and characterising differences between patients. 
			
	Although \ourmethod performs very well in practice, we see ways to improve over it. Firstly, MDL is not a magic wand. That is, while our score performs rather well in practice, we carefully constructed it to reward structure in the form of multivariate patterns. It will be interesting to see how our score and algorithms can be adapted to work directly on real-valued data. Secondly, it is worth investigating whether our current encoding can be refined, e.g.\ using prequential codes~\cite{budhathoki:15:diffnorm}. A strong point of our approach is that we allow for noise in the form of gaps in patterns. We postulate that we can further reduce the amount of redundancy in the discovered pattern sets by allowing noise \emph{in the occurrences} of a pattern, as well as when we allow \emph{overlap} between the occurrences of patterns in a cover. For both cases, however, it is not immediately clear how to adapt the score accordingly, and even more so, how to maintain the efficiency of the cover and search algorithms. 
	
	Last, but not least, we are interested in applying \ourmethod on vast time series. To accomodate, the first step would be to investigate parallelisation; the search algorithm is trivially parallelisable, as candidates can be generated and estimated in parallel, as is the covering of the data. More interesting is to investigate more efficient candidate generation schemes, in particular top-k mining, or lazy materialization of candidates. 
	
	Previous work has shown that MDL-based methods work particularly well for a wide range of data mining problems, including classification~\cite{vreeken:11:krimp,lam:14:seq} and outlier detection~\cite{smets:11:odd}. It will make for interesting future work to investigate how well \ourmethod solves such problems for multivariate event sequences. Perhaps the most promising direction of further study is that of causal inference~\cite{vreeken:15:ergo}.

\section{Conclusion} \label{sec:con}

	We studied the problem of mining interesting patterns from multivariate sequential data. We approached the problem from a pattern set mining perspective, by MDL identifying the optimal set of patterns as those that together describe the data most succinctly. We proposed the \ourmethod algorithm for efficiently discovering high-quality patterns sets from data. 

	Experiments show that \ourmethod discovers patterns planted in synthetic data with high accuracy. Moreover, it scales favourably with the length of the data, the number of attributes, and alphabet sizes. For real data, it discovers easily interpretable summaries that provide clear insight in the associations of the data. 
		
	As future work, building upon our results on the  part-of-speech tagged text data, we are collaborating with colleagues from the linguistics department to apply \ourmethod for analysis of semantically annotated text and for inferring patterns in morphologically rich languages.

\section*{Acknowledgments}
	The authors wish to thank Kathrin Grosse for her help with and analysis of the part-of-speech tagged text experiments.
	Roel Bertens and Arno Siebes are supported by the Dutch national program COMMIT.
	Jilles Vreeken is supported by the Cluster of Excellence ``Multimodal Computing and Interaction'' within the Excellence Initiative of the German Federal Government.

\balance

\bibliographystyle{abbrv}      

\bibliography{bib/abbrev,bib/references,bib/bib-jilles}

\end{document}